\documentclass{llncs}
\usepackage{url}
\usepackage{amsmath}
\usepackage{fixltx2e}
\newenvironment{keywords}{
       \list{}{\advance\topsep by0.35cm\relax\small
       \leftmargin=1cm
       \labelwidth=0.35cm
       \listparindent=0.35cm
       \itemindent\listparindent
       \rightmargin\leftmargin}\item[\hskip\labelsep
                                     \bfseries Keywords:]}
     {\endlist}
\begin{document}
%
%
\pagestyle{headings}  
%
%
\mainmatter              
\title{Towards OWL-based Knowledge Representation in Petrology}
\titlerunning{OWL in Petrology}  

\author{Alex Shkotin \and Vladimir Ryakhovsky \and Dmitry Kudryavtsev}

\authorrunning{Alex Shkotin et al.} 

\tocauthor{Alex Shkotin, Vladimir Ryakhovsky, Dmitry Kudryavtsev}
\institute{Vernadsky State Geological Museum, Russian Academy of Sciences \\
Mokhovaya 11/11, 125009, Moscow, Russia\\
\email{ashkotin@acm.org}}

\maketitle

\begin{abstract}
This paper presents our work on development of OWL-driven systems for formal representation and reasoning about terminological knowledge and facts in petrology. The long-term aim of our project is to provide solid foundations for a large-scale integration of various kinds of knowledge, including basic terms, rock classification algorithms, findings and reports. We describe three steps we have taken towards that goal here. First, we develop a semi-automated procedure for transforming a database of igneous rock samples to texts in a controlled natural language (CNL), and then a collection of OWL ontologies. Second, we create an OWL ontology of important petrology terms currently described in natural language thesauri. We describe a prototype of a tool for collecting definitions from domain experts. Third, we present an approach to formalization of current industrial standards for classification of rock samples, which requires linear equations in OWL 2. In conclusion, we discuss a range of opportunities arising from the use of semantic technologies in petrology and outline the future work in this area. 
\end{abstract}

\section{Introduction}
Petrology, a branch of geology studying rocks and their formation, plays an important role in describing Earth's crust structure, which is essential for revealing patterns in distribution of mineral resources. Similar to other natural sciences, a wealth of knowledge requiring a proper management (especially with regard to consistency) and integration has been accumulated in petrology. These tasks could be approached more efficiently, if the knowledge had been machine processable, in particular, if a \textit{formal theory of petrology} (i.e. a system of axioms, definitions and theorems \cite{I2ML}, p.33) had been available. Ontologies, especially OWL ontologies, are well suited for playing the role of a cornerstone of such theory, as they have been remarkably successful in other sciences, e.g., bioinformatics, chemistry, and health care. 

This paper describes our steps towards developing a formal theory of petrology. We focus on identifying basic terms, providing definitions to other commonly used terms i.e., terms used in industrial standards, and namely, rock types such as rhyolite or harzburgite, and formalizing the basic set of axioms. We use OWL as a main formalization tool enabling us, in particular, to automatically check our representation for consistency. 

It is only natural to start developing a theory by identifying the important terms to be later used for representing facts, e.g., knowledge about specific rock samples. Such facts are typically stored in relational databases in modern petrology, so relational databases can be used as a source of terms. We describe the conversion of one such database, namely Proba \cite{BDP} (\textit{Sample} in Russian), to a collection of OWL ontologies containing facts expressed using an initial set of currently undefined terms in the \ref{sec:db} section. 

Once the terms have been identified, we proceed to their formalization, i.e., writing their definitions in OWL. First, it is essential to define the basic terms, which can be used to define all other terms. Currently available definitions are usually stored in a semi-structured form in natural language thesauri. Besides other issues, this often leads to contradictions, especially given differences between schools in petrology. We use one such thesaurus, namely the Glossary of Igneous Rocks \cite{pgk}, to define petrological terms and relationships in an OWL ontology. In addition, we develop a webProtege-based tool to enable domain experts to work collaboratively on term definitions, in particular, to agree upon them. See the \ref{sec:glossary} section for details. 

Finally, we complement the ontology by using another rich source of term definitions - internationally adopted scientific recommendations describing rock sample classification methodologies, e.g. \textit{Igneous Rocks: A Classification and Glossary of Terms} \cite{IRCGT}. The \ref{sec:alg} section describes an approach to extracting definitions from the standard and expressing them as OWL axioms. As it stands, OWL 2 is insufficient for a complete capture of terms semantics (as specified in the standard), but this would be possible if \textit{path free linear equations} were adopted.\footnote{The details of the proposed extensions are available at \url{http://www.w3.org/TR/owl2-dr-linear/}. Our work on petrology may be viewed as a use case for supporting linear equations in future OWL versions.} We conclude the paper by summarizing our experience from the described work and outlining plans for the future.
\section{Formalizing Facts: From Database to OWL}\label{sec:db}

A considerable amount of important information is saved in databases, but in the form of data, which, unfortunately, is not a knowledge and requires an essential and laborious processing to obtain knowledge. This section describes a direct way of getting knowledge from the data: database conversion to the traditional form of knowledge, i.e. knowledge in a natural language. The natural language is limited to CNL to make this knowledge machine processable. We follow \textit{T. Kuhn: CNLs are subsets of natural languages that are restricted in a way that allows their automatic translation into formal logic.} p.5 \cite{kuhn2010doctoralthesis}. We consider CNL as a universal tool for representing a formal ontological knowledge. 

\subsubsection{The original database.} 
Proba DB \cite{BDP} contains data from 1,174 scientific articles (Bibliography table) about 49,285 samples of igneous rocks (Measurements table). Samples are collected all over the globe, which is reflected in the Localities, llocal, lglobal and lgroup tables. The samples are assigned a rock type (Rocks table), a genesis type (Errupttypes table), age (ages table), and, which is the main thing, weight percentage (Concentrations table) of chemical substances and isotopes (list in the Elements table). 

This brief description alone already shows that table and column identifiers can only approximately match the terms used by petrologists to exchange sample data. The transition to CNL also solves the problem of converting the data saved in RDB to knowledge in a form directly understandable to experts in the subject domain. 

\subsubsection{CNL sentences.} 
List 1 includes examples of all types of CNL sentence required to present all facts contained in the Proba DB. Local (internal) proper names required to name various objects within the knowledge base are used in the sentences. So, PUB5633 is the name of article number 5633 (from bibliography.id) in the DB. SAM32994 is the name of sample number 32994 (from measurements.id) in the DB, etc. Words are connected by letter ``$\_$" in compound terms. 
The text also contains well-known global proper names, for example, Iceland, Atlantic$\_$Ocean. 
\medskip

\noindent
{\it List 1. Example of CNL sentences.}
\begin{verbatim}
PUB5633 is a publication.
A title of PUB5633 is "A CONTRIBUTION TO THE GEOLOGY OF THE K...".
SAM32994 is a sample. SAM32994 is a rhyolite.
PUB5633 describes SAM32994.
PLC32994 is a place. PLC32994 is a part of Iceland.
A gathering_place of SAM32994 is PLC32994.
SUB469812 is a substance. SAM32994 includes SUB469812.
WPC469812 is a weight_percent. A value of WPC469812 is 73.95.
A component of WPC469812 is SUB469812.
\end{verbatim}

The sentence structure is very simple. A very limited natural language is actually required to record all facts contained in a RDB if RDB  is normalized. But RDB Proba is normalized not everywhere. 
Completing normalization is one of the tasks of reorganizing a DB to enable automatic conversion to knowledge. 
Rules of mapping the RDB content to CNL have been developed. These rules are the specification for SQL-scripts dumping RDB to CNL text \cite{otch08}. 

\subsubsection{OWL ontology: getting and analysis.} 
All generated sentences are ACE language \cite{fuchs99ace3Manual} sentences, and are selected so that a concrete APE compiler \footnote{Attempto Parsing Engine \url{http://attempto.ifi.uzh.ch/site/tools/}} could compile them to OWL. 
A portion of the knowledge contained in each article is separated as a text (ACE file) to be converted to an independent ontology (DL species is AL(D)). Thus, the DB will be converted to 1,174 ontologies. Columns values mainly form attribute values, but also class names (rhyolite, harzburgite) and individual names (Iceland). 
Let's consider the ontology obtained for an article with a DB number of 5633. 
The obtained classes, properties and individuals are listed below. 

\noindent 
\textit{Classes}: 
place, publication, \textbf{rhyolite}, sample, substance, weight$\_$percent. 

\noindent 
\textit{Object properties}: 
component, describes, gathering$\_$place, includes, mixture, part. 

\noindent 
\textit{Data properties}: 
authorial$\_$number, chemical$\_$formula, first$\_$page, journal$\_$reference, last$\_$page, latitude, longitude, reference, title, value, year. 

\noindent 
\textit{Individuails}: 
Atlantic$\_$Ocean, Iceland etc. 

All the terms used except rhyolite refer to contexts outside of petrology and even geology. These are the contexts of geography (place, etc), scientific publications (publication etc), solid state physics (sample, substance, weight$\_$percent etc), chemistry (chemical$\_$formula). The rest of the report focuses on obtaining rock type definitions, including that for rhyolite. 
\section{Formalizing Terminology: From Natural Language to OWL}\label{sec:glossary}

The ontology of the facts specifies that the part of names used for classes, relations, individuals belongs to a different ontology (vocabulary). 
This dictionary ontology is supposed to provide term definitions, and the author of the article has exactly this understanding in mind. 
Such scientific terms are normally already collected in a dictionary, for example, Petrographic Dictionary \cite{pgs}, Dictionary of Geological Terms \cite{sgt}, Dictionary of Igneous Rocks Terms \cite{pgk}, Glossary of Geology \cite{GoG}. 
The dictionary represents a very important and specific type of knowledge. It is based on subject domain terms and informal definitions of these terms. 
Example: harzburgite rock type article from \cite{IRCGT}, p.88: 
\begin{quote} 
HARZBURGITE. An ultramafic plutonic rock composed essentially of olivine and orthopyroxene. Now defined modally in the ultramafic rock classification (Fig. 2.9, p.28). 
(Rosenbusch, 1887, p.269; Harzburg, Harz Mts, Lower Saxony, Germany; Troeger 732; 
Johannsen v.4, p.438; Tomkeieff p.247) 
\end{quote} 

We have converted a specific dictionary (\cite{pgk}) initially presented by authors as an html page to an OWL ontology. 
%
%
%
We begin the formalization of relations between terms (for example, synonymy) and term properties (for example, become outdated). 

\subsubsection{Converting the dictionary text to ontology.} 
We took the Dictionary of Terms of Igneous Rock Types compiled by the Interdepartmental Petrographic Committee in the Department of Earth Sciences of the Russian Academy of Sciences \cite{pgk}. The dictionary contains 1,567 articles, the overwhelming majority of them being rock names. The dictionary structure and conversion procedures required to get the ontology are described in \cite{otch09} and most important below. 

\noindent 
\textit{Vocabulary}: 
 Words are connected by letter ``$\_$" in compound terms. 

\noindent 
\textit{Article title}: 
The dictionary article title contains a Russian term and its English equivalent in a simple case, but its both Russian and English synonyms are often specified as well. 
Each term present in the title generates an ontology class. 
Thus, the ontology will contain classes in Russian and in English. 
All terms from one title are considered synonyms, i.e. their classes are declared equivalent. 
These conversions resulted in 3,179 classes and 1,659 class equivalence axioms having appeared in the ontology. 

\noindent 
\textit{The text of the article}: 
The basic dictionary article text parts are: term definition, comment, list of links to references (normally at the end), term origin description (normally located on the list of references after the article, in which the term was introduced).
Comments and a list of links to references located in some parts of the ontology in the form of separate annotations are supposed to be selected from the text of the article. 

The dictionary ontology (DL species is ALUF(D)) is published\footnote{\url{http://earth.jscc.ru/ontologies/dic.owl}} and can be viewed using any ontology browser at this moment. 

\subsubsection{Collective management of scientific term definitions.} 
Another copy of the ontology is accessible by means of webProtege \footnote{\url{http://protegewiki.stanford.edu/index.php/WebProtege}} installed on the Geology portal\footnote{\url{http://earth.jscc.ru/webprotege/}}. 
The dictionary ontology is 'dic' there. 

It is important that a prefix and a namespace be assigned to each dictionary. We have for terms of the ontology itself, terms from the Moscow State University Geoweb portal, terms from the Petrographic Code of Russia \cite{PgKR}, and terms from the \cite{pgk} dictionary, respectively: 
\begin{verbatim}
prefix dic: <//earth.jscc.ru/ontologies/dic.owl#>
prefix gwr: <//wiki.web.ru/wiki#>
prefix pgcc: <//www.igem.ru/site/petrokomitet/code#>
prefix pgc: <//www.igem.ru/site/petrokomitet/slovar#>
\end{verbatim}

A formal term meaning definition is critical for developing a formal theory. 
For example, the current version of the dictionary provides a formal definition of the abessedite rock type (see Portlet Axioms for dic:abessedite), and namely 
\newline
\newline
\texttt{peridotite and minerals\_mixture and} \\
\texttt{ contains\_mineral only (olivin or hornblende or phlogopite)}
\newline

This formula is written using the Manchester OWL syntax. It is important that petrologists are able to read it. 
The process of obtaining a formal (mathematical) definition, especially in a form clear to experts, is described further, and is one of project's main ultimate goals. 
The \cite{otch09} report contains details of the work done. 
\section{Formalizing Rock Classification}\label{sec:alg}
Rules of rock type assignment to samples are described in \cite{IRCGT} and consist of a description of initial-classification algorithm and diagrams of final classification by percentage of essential minerals. 
We begin with a specification of all parts of the algorithm, sample data being its input and term (word combination) representing sample rock type its output. The algorithm is written as a set of functions in the form of a flowchart clear to petrologists. 

The algorithm uses some real-valued functions and unary predicates. These functions and predicates are supposed to have value on any solid \cite{RCK}. Some of these functions and predicates have been given definitions, definitions should be found for other ones, and some will probably remain without definitions and will enter in the formal theory as primary ones. The algorithm and necessary definitions are given for ultramafic types of plutonic rock as an example. It is shown then how to get formal definitions of some types of rock from the algorithm. 

\texttt{VPC} means mineral Volume Percentage Content of the sample and is also known as ``volume modal data". 

We name an algorithm function (for example, \texttt{ultramafic\_rock\_type}) receiving sample data at its input and returning a sample rock type name \textit{classifying}. 

\subsubsection{Quantitative and Qualitative Characteristics.} 
We need unary real-valued functions returning the volume percentage of minerals in a solid. The full set of minerals required for the algorithm will be gradually clarified. 

The following functions of one argument returning a real number were required till now: 
\texttt{VPC\_melilite}, \texttt{VPC\_kalsilite}, \texttt{VPC\_leucite}, \texttt{VPC\_Ol}, \texttt{VPC\_Opx}, \texttt{VPC\_Cpx}, \texttt{VPC\_hornblende}, \texttt{VPC\_garnet}, \texttt{VPC\_spinel}, and \texttt{VPC\_biotite}. 
These functions are primary and may be measured. 

We also need the VPC of groups of minerals (see \cite{IRCGT} p. 4, \cite{BGSRCS} p. 6): 
\texttt{VPC\_Q}, \texttt{VPC\_A}, \texttt{VPC\_P}, \texttt{VPC\_F} and \texttt{VPC\_M}. 
It is clear that these functions have definitions. The \texttt{VPC\_M} definition is given below. 

The following unary predicates will be required to describe the sample: 
\texttt{pyroclastic}, \texttt{kimberlite}, \texttt{lamproite}, \texttt{lamprophyre}, \texttt{charnockite}, \texttt{plutonic}, and \texttt{volcanic}. 
All of these predicates are supposed to have definitions. The definition of \texttt{pyroclastic} is given below. 
\subsubsection{Definitions.} 
All the definitions currently available can be found in a technical report \cite{otch10}. 
We show typical examples here. All definitions are based on two sources: ``Igneous Rocks: A Classification and Glossary of Terms'' \cite{IRCGT} and `BGS Rock Classification Scheme'' \cite{BGSRCS}, and are confirmed by petrologists. 

\noindent \texttt{VPC\_Px:} the modal content of pyroxenes (required to classify some plutonic rocks):
\begin{align*}
\texttt{VPC\_Px(x) =} \textsubscript{def} \texttt{ VPC\_Opx(x)} + \texttt{VPC\_Cpx(x)}
\end{align*} 
Where \texttt{ =}\textsubscript{def} means by definition\cite{ddm}.

\noindent \texttt{VPC\_OOC} and \texttt{VPC\_OPH:} VPC of mineral groups. 
We need these definitions to formalize the diagrams on Fig. 2.9, p. 28 of \cite{IRCGT}. 
\begin{align*}
\texttt{VPC\_OOC(x) =} & \textsubscript{def}  \texttt{ VPC\_Ol(x)} + \texttt{VPC\_Opx(x)} + \texttt{VPC\_Cpx(x)} \\
\texttt{VPC\_OPH(x) =} & \textsubscript{def}  \texttt{ VPC\_Ol(x)} + \texttt{VPC\_Px(x)} + \texttt{VPC\_hornblende(x)} 
\end{align*}

\noindent \texttt{VPC\_M:} returns volume percentage of group M (mafic) minerals in the sample (p. 4, 28 see \cite{IRCGT}, and especially \cite{BGSRCS} p. 6). 
Following the direct instructions given in \cite{BGSRCS} p. 6: 
\begin{quote} 
\textit{M = mafic and related minerals, that is all other minerals apart from QAPF;...} 
\end{quote}
we obtain the definition: 
\begin{align*}
\texttt{VPC\_M(x) =}  \textsubscript{def}\texttt{  100 - (VPC\_Q(x)} + \texttt{VPC\_A(x)} + \texttt{VPC\_P(x)} + \texttt{VPC\_F(x)}) 
\end{align*} 

\noindent \texttt{pyroclastic:}
We mainly rely on the 2.2 PYROCLASTIC ROCKS AND TEPHRA section \cite{IRCGT}, p. 7. 
\begin{align*}
\texttt{pyroclastic(x) =}  \textsubscript{def}  	\texttt{ clastic(x)} \land (&\forall \texttt{y clast(y)} \land \texttt{part\_of(y,x)} \\
&\rightarrow \texttt{volcanic\_eruption\_result(y)}) 
\end{align*} 
This can also be represented in DL:
\begin{align*}
\texttt{pyroclastic} \equiv  \texttt{clastic} \sqcap \forall \texttt{(part\_of} \circ \texttt{id(clast))} ^-. \texttt{volcanic\_eruption\_result}
\end{align*} 

\subsubsection{Algorithm.} 
Our algorithm is a further formalization (and elaboration!) of the classification rules provided in the \cite{IRCGT}. The algorithm is written as a set of function flowcharts, the main function being the classifying rock$\_$type function. This function should be invoked to classify a sample. 
We have also created flowcharts for the ultramafic rock classifying function and two diagrams on Fig.2.9 \cite{IRCGT}, p. 28: OOC$\_$diagram$\_$field (the upper triangle) and OPH$\_$diagram$\_$field (the lower triangle). 
The IUGS diagram flowcharts are deliberately presented as a chain of if-nodes, each one being responsible for one specific diagram area. Each if-condition represents a system of linear inequalities. The set of such conditions has important mathematical properties: 
\begin{itemize}
\item Any two conditions are incompatible, since areas corresponding to them are mutually disjoint 
\item The union of all conditions gives inequalities for a triangle, since conditions cover the entire triangle
\end{itemize}
It is important that the described properties can be checked automatically if definitions are loaded in a reasoner  working with linear inequalities. 
\subsubsection{Rock type predicate definition.} 
The classification algorithm implicitly contains definitions of all types of igneous rock. Definitions can be obtained from the algorithm in the form of formulas one free variable formulas of predicate calculus of first order with numbers. The formula structure shows the complexity of the concept behind the term, and also specifies all the concepts underlying a term. This is extremely important for finding the primary concepts. 
We have quite formally, i.e. using mathematical conversions, obtained formulas for the harzburgite and dunite predicates. 

\texttt{harzburgite:} 
when applied to the sample, the harzburgite predicate should give ``true" if the sample is harzburgite, and ``false" otherwise. 
Flowcharts have to be tracked from top to bottom, and conditions leading to a OOC$\_$diagram$\_$field flowchart node producing the ``harzburgite" value collected, to get a predicate. These conditions should be connected by the logical operation ``and". 
The conversions will give the following formula: 
\newline
\newline
\noindent
\texttt{harzburgite(x) =} \textsubscript{def}  \texttt{ plutonic(x)} $\land$ $\lnot$ \texttt{(pyroclastic(x)} $\lor$ \texttt{kimberlite(x)}

$\lor$ \texttt{lamproite(x)} $\lor$ \texttt{lamprophyre(x)} $\lor$ \texttt{charnockite(x))}

$\land$ \texttt{VPC\_carbonates(x)}$\leq$ \texttt{50} $\land$ \texttt{VPC\_melilite(x)}$\leq$ 10 $\land$ \texttt{VPC\_M(x)} $\geq$ 90 

$\land$ \texttt{VPC\_kalsilite(x)=0} $\land$ \texttt{VPC\_leucite(x)=0} $\land$ \texttt{VPC\_hornblende(x)=0}

$\land$ \texttt{0.4*VPC\_OOC(x)}$\leq$ \texttt{VPC\_Ol(x)}$\leq$ \texttt{0.9*VPC\_OOC(x)}

$\land$ \texttt{VPC\_Cpx(x)<0.05*VPC\_OOC(x)} 
\newline
\newline
Thus, a precise definition of the harzburgite igneous rock type consists of three parts: 
\begin{enumerate}
\item Qualitative characteristics (lines 1, 2). 
\item Absolute restrictions on modal data (lines 3, 4). 
\item  Relative restrictions on modal data (lines 5, 6).
\end{enumerate}

Now we can compare this definition with the informal definition quoted in Section 3: the formal definition is more complete. It does not suppose anything and does not refer to the diagram. It contains the necessary part of the diagram. 

\section{Lessons Learnt, What is Next?} 

This paper describes our experience of converting the petrological information stored in databases, glossaries, and classification standards to a formal OWL-based representation. A similar approach, i.e. one based on providing unambiguous and consistent definitions for all terms, can be used in developing a formal theory for virtually any scientific area. We will now briefly summarize the results and outline plans for the future. 

\noindent 
\textit{From data to knowledge}. Moving from a database of petrological facts to a knowledge base is beneficial from multiple perspectives. Firstly, the new representation is richer and enables generation of sentences in a controlled natural language, which, in our experience, are understandable to geologists. They can be used not only as an interface to the KB, but also to annotate publications, which should lead to increased amounts of machine-processable metadata. Secondly, the KB (equipped with a CNL-based interface and a SPARQL endpoint) can be integrated with the ontology that provides the vocabulary. This is important for ensuring a consistent use of the terminology across all information systems using the KB. The stored knowledge can be further integrated with other available datasets, e.g. those provided by the EarthChem consortium. \footnote{EarthChem is a community-driven effort to facilitate the preservation, discovery and visualization of and access to the broadest and richest geochemical datasets possible: \url{http://www.earthchem.org}.} 

\noindent 
\textit{Centralized vocabulary}. Providing a controlled vocabulary is essential for managing the knowledge. In our case, it was most important to collect the terms used in the database in a single OWL ontology, and give them unambiguous definitions along with human-readable annotations. This is a substantial improvement compared to the previous situation where terms were defined informally and in multiple, often contradictory sources. The resulting system can be used both as a dictionary (for people and applications i.e., via SPARQL) and as a tool for collaborative work on terminology. 

\noindent 
\textit{Rock classification}. The formal definitions of the terms captured in standard OWL are not detailed enough to support automated rock sample classification, which is one of the most important use cases in petrology. To this end, we have investigated the possibility of complementing the definitions with quantitative restrictions on their mineral composition. Such restrictions can be defined using linear equations, a possible extension to the current data ranges in OWL 2.

Similarly to databases and glossaries, the classification recommendations, namely \cite{IRCGT}, are sometimes ambiguous and incomplete as well, so their formalization requires collaboration with petrologists from the Subcommission on the Systematics of Igneous Rocks of the International Union of Geological Sciences. However, we managed to identify some predicates and functions requiring definitions, which can be used as building blocks of a formal theory. Following the methodology described in the \ref{sec:alg} section, we have obtained detailed definitions for two types of rock as well as for some auxiliary terms. We plan to extend this work to cover all rock types in the classification. 

Our work enables answering questions like Is a current object a sample of a certain rock? by performing instance checking, a standard reasoning task in OWL. However, this can be extended to query answering to find all possible rock types for a specific sample or to find all samples of a specific type in the KB. This, however, requires reasoning with linear inequalities, which is not supported at large scale at the moment (some reasoners are available, e.g. RACER). 

Finally, we would like to stress that our approach to formalization differs from what can be seen in many biological and chemical ontologies. They are often deep class hierarchies with numerous asserted subsumptions between class names and with relatively few definitions. We focus on providing detailed definitions (using standard OWL and linear equations) instead, and plan to rely on automated reasoners to build and maintain the hierarchy. This may enable use of the ontologies in a broader range of situations as illustrated by rock sample classification. 

\subsubsection{Acknowledgments.} 

We would like to thank Dr. Kaarel Kaljurand from Attempto group for the idea of using proper names, Dr. Stephen M. Richard from Arizona Geological Survey for comments on the report \cite{otch10}, helpful discussion and reference to \cite{BGSRCS}; and Pavel Klinov from the University of Manchester for numerous invaluable comments on this paper. 
%
\bibliographystyle{splncs03}
\bibliography{petrology}

\end{document}